\documentclass[twoside]{article}

\usepackage[accepted]{aistats2024}
\usepackage[round]{natbib}

\usepackage{caption}
\usepackage{setspace}
\usepackage{graphicx}
\usepackage{xcolor}
\usepackage{booktabs}
\usepackage{amssymb}
\usepackage{amsmath}
\usepackage{multirow}
\usepackage{array}
\usepackage{url}
\usepackage{hyperref}
\usepackage[flushleft]{threeparttable}
\DeclareMathOperator{\diag}{diag}

\begin{document}

%

%

\twocolumn[

\aistatstitle{Better Batch for Deep Probabilistic Time Series Forecasting}

\aistatsauthor{ Vincent Zhihao Zheng \And Seongjin Choi \And  Lijun Sun }

\aistatsaddress{ McGill University \And  University of Minnesota \And McGill University } ]

\begin{abstract}
Deep probabilistic time series forecasting has gained attention for its ability to provide nonlinear approximation and valuable uncertainty quantification for decision-making. However, existing models often oversimplify the problem by assuming a time-independent error process and overlooking serial correlation. To overcome this limitation, we propose an innovative training method that incorporates error autocorrelation to enhance probabilistic forecasting accuracy. Our method constructs a mini-batch as a collection of $D$ consecutive time series segments for model training. It explicitly learns a time-varying covariance matrix over each mini-batch, encoding error correlation among adjacent time steps. The learned covariance matrix can be used to improve prediction accuracy and enhance uncertainty quantification. We evaluate our method on two different neural forecasting models and multiple public datasets. Experimental results confirm the effectiveness of the proposed approach in improving the performance of both models across a range of datasets, resulting in notable improvements in predictive accuracy.
\end{abstract}

\section{\MakeUppercase{Introduction}}

Time series forecasting stands as one of the primary tasks in the field of deep learning (DL) due to its broad range of applications \citep{benidis2022deep}. Essentially, the problem of time series forecasting can be classified into deterministic forecasting and probabilistic forecasting. Deterministic forecasting provides point estimates for future time series values, while probabilistic forecasting goes a step further by providing a distribution that quantifies the uncertainty associated with the predictions. As additional information on uncertainty assists users in making more informed decisions, probabilistic forecasting has become increasingly attractive and extensive efforts have been made to enhance uncertainty quantification. In time series analysis, errors can exhibit correlation for various reasons, such as the omission of essential covariates or model inadequacy. Autocorrelation (also known as serial correlation) and contemporaneous correlation are two common types of correlation in time series forecasting. Autocorrelation captures the temporal correlation present in errors, whereas contemporaneous correlation refers to the correlation among different time series at the same time. 

This paper primarily investigates the issue of autocorrelation in errors. Modeling error autocorrelation is an important field in the statistical analysis of time series. A widely adopted method for representing autocorrelated errors is assuming the error series follows an autoregressive integrated moving average (ARIMA) process \citep{hyndman2018forecasting}. Similar issues may arise in learning nonlinear DL-based forecasting models. Previous studies have attempted to model error autocorrelation in deterministic DL models using the concept of dynamic regression \citep{sun2021adjusting,zheng2023enhancing}, assuming that the errors follow a first-order autoregressive process. However, since both neural networks and correlated errors can explain the data, these models may face challenges in balancing the two sources in the absence of an overall covariance structure. More importantly, these methods are not readily applicable to probabilistic models, where the model output typically consists of parameters of the predictive distribution rather than the estimated time series values. 

In this paper, we propose a novel batch structure that allows us to explicitly model error autocorrelation. Each batch comprises multiple mini-batches, with each mini-batch grouping a fixed number of consecutive training instances. Our main idea draws inspiration from the generalized least squares (GLS) method used in linear regression models with dependent errors. We extend the Gaussian likelihood of a univariate model to a multivariate Gaussian likelihood by incorporating a time-varying covariance matrix that encodes error autocorrelation within a mini-batch. The covariance matrix is decomposed into two components, a scale vector and a correlation matrix, that are both time-varying. In particular, we parameterize the correlation matrix using a weighted sum of several base kernel matrices, and the weights are dynamically generated from the output of the base probabilistic forecasting model. This enables us to improve the accuracy of estimated distribution parameters during prediction by using the learned dynamic covariance matrix to account for previously observed residuals. By explicitly modeling dynamic error covariance, our method enhances training flexibility, improves time series prediction accuracy, and provides high-quality uncertainty quantification. Our main contributions are as follows:
\begin{itemize}
    \item We propose a novel method that enhances the training and prediction of univariate probabilistic time series models by learning a time-varying covariance matrix that captures the correlated errors within a mini-batch. 
    \item We parameterize the dynamic correlation matrix with a weighted sum of several base kernel matrices. This ensures that the correlation matrix is a positive definite symmetric matrix with unit diagonals. This approach allows us to jointly learn the dynamic weights alongside the base model.
    \item We evaluate the effectiveness of the proposed approach on two base models with distinct architectures, DeepAR and Transformer, using multiple public datasets. Our method effectively captures the autocorrelation in errors and thus offers enhanced prediction quality. Importantly, these improvements are achieved through a statistical formulation without substantially increasing the number of parameters in the model.
\end{itemize}

\section{\MakeUppercase{Preliminaries}}\label{preliminaries}

\subsection{Probabilistic Time Series Forecasting}

Denote \(\mathbf{z}_{t}=\left[z_{1,t},\dots,z_{N,t}\right]^\top \in \mathbb{R}^{N}\) the time series variables at time step $t$, where \(N\) is the number of time series. Given the observed history $\{\mathbf{z}_{t}\}_{t=1}^T$, the task of probabilistic time series forecasting involves formulating the estimation of the joint conditional distribution \(p\left(\mathbf{z}_{{T+1}:{T+Q}} \mid \mathbf{z}_{{T-P+1}:{T}}; \mathbf{x}_{{T-P+1}:{T+Q}}\right)\). Here, $\mathbf{z}_{{t_1}:{t_2}} =\left[\mathbf{z}_{t_1},\ldots,\mathbf{z}_{t_2}\right]$, and  \(\mathbf{x}_{t}\) represents known time-dependent or time-independent covariates, such as the time of day or the time series identifier. Put differently, our focus is on predicting \(Q\) future values based on \(P\) historical values and covariates. The model can be further decomposed as
\begin{multline}\label{eqn:prob1}
    p\left(\mathbf{z}_{{T+1}:{T+Q}} \mid \mathbf{z}_{{T-P+1}:{T}}; \mathbf{x}_{{T-P+1}:{T+Q}}\right) \\ =\prod_{t=T+1}^{T+Q} p\left(\mathbf{z}_{t} \mid \mathbf{z}_{{t-P}:{t-1}}; \mathbf{x}_{{t-P}:{t}}\right),
\end{multline}
which is an autoregressive model that can be used for performing multi-step-ahead forecasting in a rolling manner. In this case, samples are drawn within the prediction range (\(t\geq T+1\)) and fed back for the subsequent time step until reaching the end of the prediction range. The conditioning is usually expressed as a state vector \(\mathbf{h}_t\) of a transition dynamics \(f_{\Theta}\) that evolves over time $\mathbf{h}_t = f_{\Theta}\left(\mathbf{h}_{t-1}, \mathbf{z}_{t-1}, \mathbf{x}_{t}\right)$. Hence, Eq.~\eqref{eqn:prob1} can be expressed in a simplified form as 
\begin{multline}\label{eqn:prob2}
    p\left(\mathbf{z}_{{T+1}:{T+Q}} \mid \mathbf{z}_{{T-P+1}:{T}}; \mathbf{x}_{{T-P+1}:{T+Q}}\right) \\ =\prod_{t=T+1}^{T+Q} p\left(\mathbf{z}_{t} \mid \mathbf{h}_{t}\right),
\end{multline}
where \(\mathbf{h}_{t}\) is mapped to the parameters of a specific parametric distribution (e.g., Gaussian, Poisson). When $N=1$, the problem is reduced to a univariate model:
\begin{multline}\label{eqn:prob2_uni}
    p\left(\mathbf{z}_{i,T+1:T+Q} \mid \mathbf{z}_{i,T-P+1:T}; \mathbf{x}_{i,T-P+1:T+Q}\right) \\ =\prod_{t=T+1}^{T+Q} p\left(z_{i,t} \mid \mathbf{h}_{i,t}\right),
\end{multline}
where \(i\) is the identifier of a time series.

\subsection{Error Autocorrelation}

In the majority of probabilistic time series forecasting literature, the data under consideration is typically continuous, and the errors are assumed to follow an independent Gaussian distribution. Consequently, the time series variable associated with this framework is expected to follow a Gaussian distribution:
\begin{equation}\label{eqn:z_norm}
    \left.z_{i,t} \mid \mathbf{h}_{i,t}\right.\sim\mathcal{N}\left( \mu(\mathbf{h}_{i,t}), \sigma^2(\mathbf{h}_{i,t})\right),
\end{equation}
where $\mu(\cdot)$ and $\sigma(\cdot)$ map the state vector $\mathbf{h}_{i,t}$ to the mean and standard deviation of a Gaussian distribution. For instance, the DeepAR model \citep{salinas2020deepar} adopts $\mu(\mathbf{h}_{i,t})=\mathbf{w}_{\mu}^\top \mathbf{h}_{i,t}+b_{\mu}$ and $\sigma(\mathbf{h}_{i,t})=\log(1+\exp(\mathbf{w}_{\sigma}^\top \mathbf{h}_{i,t}+b_{\sigma}))$ and both parameters are time-varying. With Eq.~\eqref{eqn:z_norm}, we can decompose the time series variable into $z_{i,t}=\mu_{i,t}+\eta_{i,t}$, where $\eta_{i,t} \sim \mathcal{N}(0,\sigma_{i,t}^2)$. Assuming the errors to be independent corresponds to $\operatorname{Cov}(\eta_{i,t-\Delta}, \eta_{i,t})=0$, $ \forall \Delta \neq 0.$ In the following of this paper, we focus on this setting with Gaussian errors. 

\begin{figure}[!t]
  \centering
  \includegraphics[width=0.4\textwidth, interpolate=false]{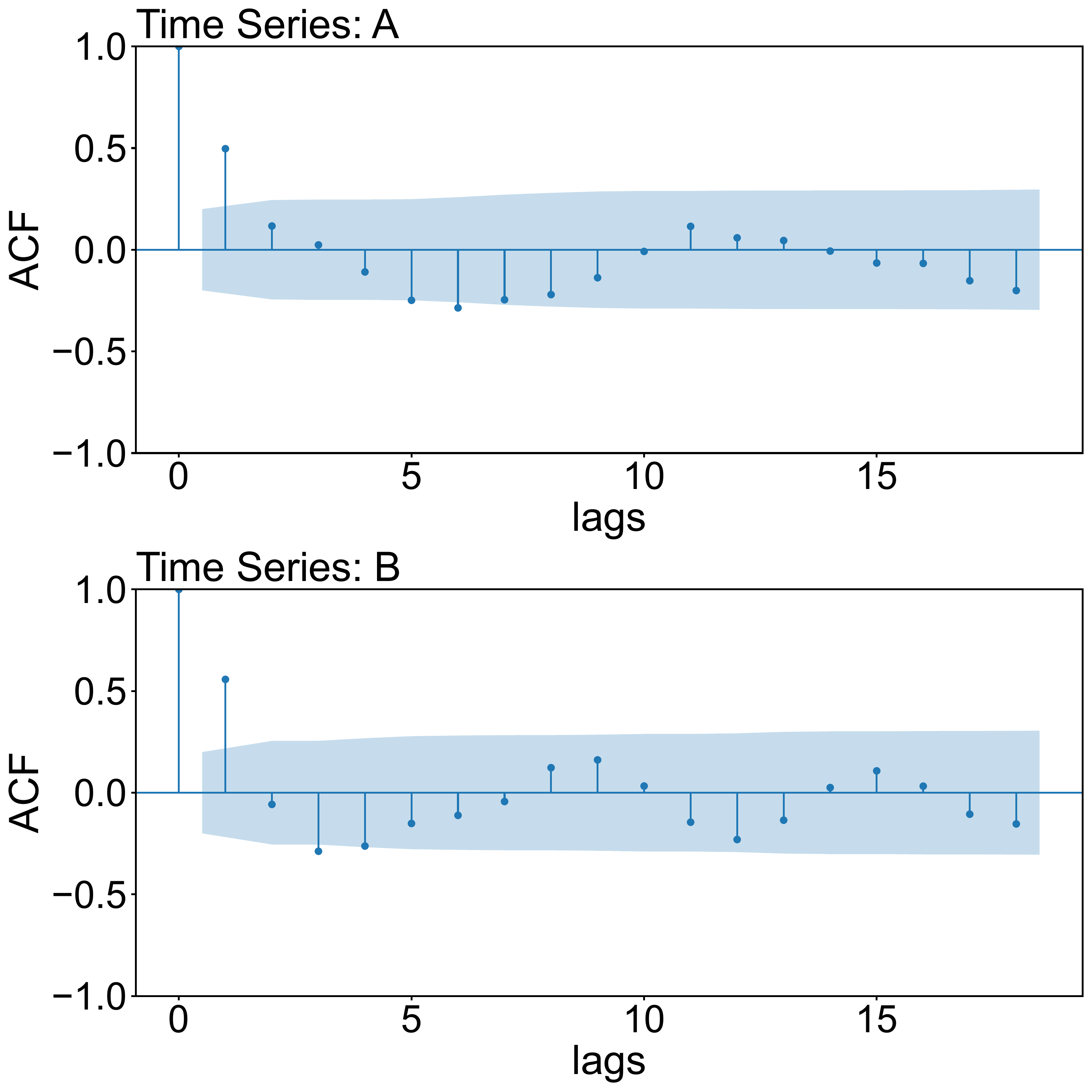}
  \caption{Autocorrelation Function (ACF) of the One-step-ahead Prediction Residuals. The results depict the prediction outcomes generated by DeepAR for two time series in the $\mathtt{m4\_hourly}$ dataset. The shaded area indicates the 95\% confidence interval, highlighting regions where the correlation is statistically insignificant.}
\label{fig:acf_exp}
\end{figure}

When there exists serial correlation in the error process, we will have \(\boldsymbol{\eta}_{T+1:T +Q}=\left[\eta_{i,T+1},\ldots,\eta_{i,T+Q}\right]^{\top}\) follows a multivariate Gaussian distribution \(\mathcal{N}(\boldsymbol{0},  \boldsymbol{\Sigma})\), where \( \boldsymbol{\Sigma}\) is the covariance matrix. Fig.~\ref{fig:acf_exp} gives an example of residual autocorrelation functions (ACF) calculated using the prediction results of DeepAR on two time series from the $\mathtt{m4\_hourly}$ dataset. The plot reveals a prevalent issue of lag-1 autocorrelation. Ignoring the systematic autocorrelation will undermine the performance of forecasting.

\section{\MakeUppercase{Related Work}}\label{literatures}

\subsection{Probabilistic Time Series Forecasting}

Probabilistic forecasting aims to offer the predictive distribution of the target variable rather than producing a single-point estimate, as seen in deterministic forecasting. Essentially, there are two approaches to achieve this: via the probability density function (PDF) or through the quantile function \citep{benidis2022deep}. For example, MQ-RNN \citep{wen2017multi} employs sequence-to-sequence (Seq2Seq) recurrent neural networks (RNNs) to directly output specific quantiles of the predictive distribution. 

In contrast, PDF-based models typically use a probabilistic model to describe the distribution of the target variables, with neural networks often employed to generate the parameters of this probabilistic model. For example, DeepAR \citep{salinas2020deepar} employs RNNs to model the transitions of hidden state. The hidden state at each time step is used to generate the parameters of a Gaussian distribution. Consequently, prediction samples for the target variables can be drawn from this distribution. GPVar \citep{salinas2019high}, a multivariate extension of DeepAR, utilizes a Gaussian copula to transform the original observations into Gaussian variables. Subsequently, a multivariate Gaussian distribution is assumed for these transformed variables. State space model (SSM) is also a popular choice for the probabilistic model. For instance, the deep SSM model proposed by \cite{rangapuram2018deep} employs RNNs to generate parameters for the state space model, facilitating the generation of prediction samples. The normalizing kalman filter \citep{de2020normalizing} combines normalizing flows with the linear Gaussian state space model. This integration enables the modeling of nonlinear dynamics using RNNs and a more flexible probability density function for observations with normalizing flows. The deep factor model \citep{wang2019deep} employs a deterministic model and a probabilistic model separately to capture the global and local random effects of time series. The probabilistic model could be any classical probabilistic time series model such as a Gaussian noise process. Conversely, the global model, parameterized by neural networks, is dedicated to representing the deterministic patterns inherent in time series. 

Various efforts have been undertaken to improve the quality of probabilistic forecasting. One avenue involves enhancing expressive conditioning for probabilistic models. For example, some approaches involve replacing RNNs with Transformer to model latent state dynamics, thus mitigating the Markovian assumption inherent in RNNs \citep{tang2021probabilistic}. Another approach focuses on adopting more sophisticated distribution forms, such as normalizing flows \citep{rasul2020multivariate} and diffusion models \citep{rasul2021autoregressive}. For a recent and comprehensive review, readers are referred to \cite{benidis2022deep}.

\subsection{Modeling Correlated Errors}

Error correlation can be categorized into two main types: autocorrelation and contemporaneous correlation. In time series analysis, autocorrelation occurs when errors in a time series are correlated over different time points, while contemporaneous correlation refers to the correlation between errors at the same time step. 

Autocorrelation has been extensively studied in classical time series models \citep{prado2021time,hyndman2018forecasting,hamilton2020time}. Statistical frameworks, including autoregressive (AR) and moving average (MA) models, have been well-developed to address autocorrelation. One important method is dynamic regression \citep{hyndman2018forecasting}, where errors are assumed to follow an ARIMA process. Recent DL-based models have also attempted to handle autocorrelation, such as re-parameterizing the input and output of neural networks to model first-order error autocorrelation with an AR process \citep{sun2021adjusting}. This method enhances the performance of DL-based time series models, and the parameters introduced to model autocorrelation can be jointly optimized with the base DL model. However, it is limited to one-step-ahead forecasting. This method was later extended to multivariate models for Seq2Seq traffic forecasting tasks by \citet{zheng2023enhancing}, assuming a matrix AR process for the matrix-valued errors. 

Contemporaneous correlation modeling often appears in spatial regression tasks. For instance, \citet{jia2020residual} proposed using a multivariate Gaussian distribution to model label correlation in the node regression problem, where the predicted label at each node is often considered conditionally independent. This correlation can later be utilized to refine predictions for unknown nodes using information from known node labels. The introduction of GLS loss in \citet{zhan2023neural} captures the spatial correlation of errors in neural networks for geospatial data, bridging deep learning with Gaussian processes. A similar approach, involving the utilization of GLS loss with random forest, was proposed by \cite{saha2023random}. Contemporaneous correlation is also modeled in time series forecasting to capture the interdependence between time series. The methods range from using a parametric multivariate Gaussian distribution \citep{salinas2019high} to more expressive generative models such as normalizing flows \citep{rasul2020multivariate} and diffusion models \citep{rasul2021autoregressive}. In \citet{choi2022spatiotemporal}, a dynamic mixture of matrix normal distributions was proposed to characterize spatiotemporally correlated errors in multivariate Seq2Seq traffic forecasting tasks.

To the best of our knowledge, our study presents an innovative training approach to addressing error autocorrelation in probabilistic time series forecasting. Our proposed method is closely related to those introduced in \cite{sun2021adjusting}, \cite{zhan2023neural}, and \cite{saha2023random}. While \cite{zhan2023neural}, and \cite{saha2023random} primarily concentrate on modeling contemporaneous correlation, and \cite{sun2021adjusting} focuses on modeling autocorrelation in deterministic forecasting. Our method concentrates on learning temporally correlated errors in the probabilistic time series forecasting context. We utilize a dynamic covariance matrix to capture autocorrelation within a mini-batch, which is simultaneously learned alongside the base model. The introduction of the error covariance matrix not only provides a statistical framework for characterizing error autocorrelation but also enhances prediction accuracy.

\section{\MakeUppercase{Our Method}}\label{methods}

Our approach is based on the formulation presented in Eq.~\eqref{eqn:prob2_uni}, using an autoregressive model as the base model. Given the primary focus of this paper on univariate models, we will omit the subscript $i$ for the remainder of this paper. In a general sense, an autoregressive probabilistic forecasting model comprises two key components: firstly, a transition model (e.g., an RNNs) to characterize state transitions $\mathbf{h}_t$, and secondly, a distribution head $\theta$ responsible for mapping $\mathbf{h}_t$ to the parameters governing the desired distribution. Furthermore, the encoder-decoder framework is employed to facilitate multi-step forecasting, wherein an input sequence spanning $P$ time steps is used to generate an output sequence spanning $Q$ time steps. The likelihood is expressed as $p\left(z_{t} \mid \theta(\mathbf{h}_{t})\right)$ for an individual observation, and in the case of employing a Gaussian distribution, $\theta(\mathbf{h}_{t})$ takes the form of $(\mu_{t}, \sigma_{t})$. In the training batch, the target time series variable can be decomposed as
\begin{equation}
    z_{t} = \mu_{t} + \sigma_{t}\epsilon_{t}, 
\end{equation}
where $\epsilon_{t}$ is the normalized error term, which usually follows the assumption that $\epsilon_{t}=\frac{z_{t}-\mu_{t}}{\sigma_{t}} \overset{\text{iid}}{\sim} \mathcal{N}(0,1)$. This assumption implies that $\epsilon_{t}$ are independent and identically distributed according to a standard normal distribution. Consequently, the parameters of the model can be optimized by maximizing its log-likelihood: 
\begin{equation}\label{eqn:nll_uni}
\mathcal{L}=\sum_{t=1}^{T} \log p\left(z_{t} \mid \theta\left(\mathbf{h}_{t}\right)\right)\propto \sum_{t=1}^{T} -\frac{1}{2}\epsilon_{t}^2-\ln \sigma_{t}.
\end{equation}
We adopt a unified univariate model trained across all time series, rather than training individual models for each time series. Moreover, if we assume the error process to be isotropic, the loss function is equivalent to the Mean Squared Error (MSE) commonly employed in training deterministic models \citep{sun2021adjusting}. However, this assumption of independence ignores the potential serial correlation in $\epsilon_{t}$.

\subsection{Training with Mini-batch}

\begin{figure*}[!t]
  \centering
  \includegraphics[width=0.7\textwidth, interpolate=false]{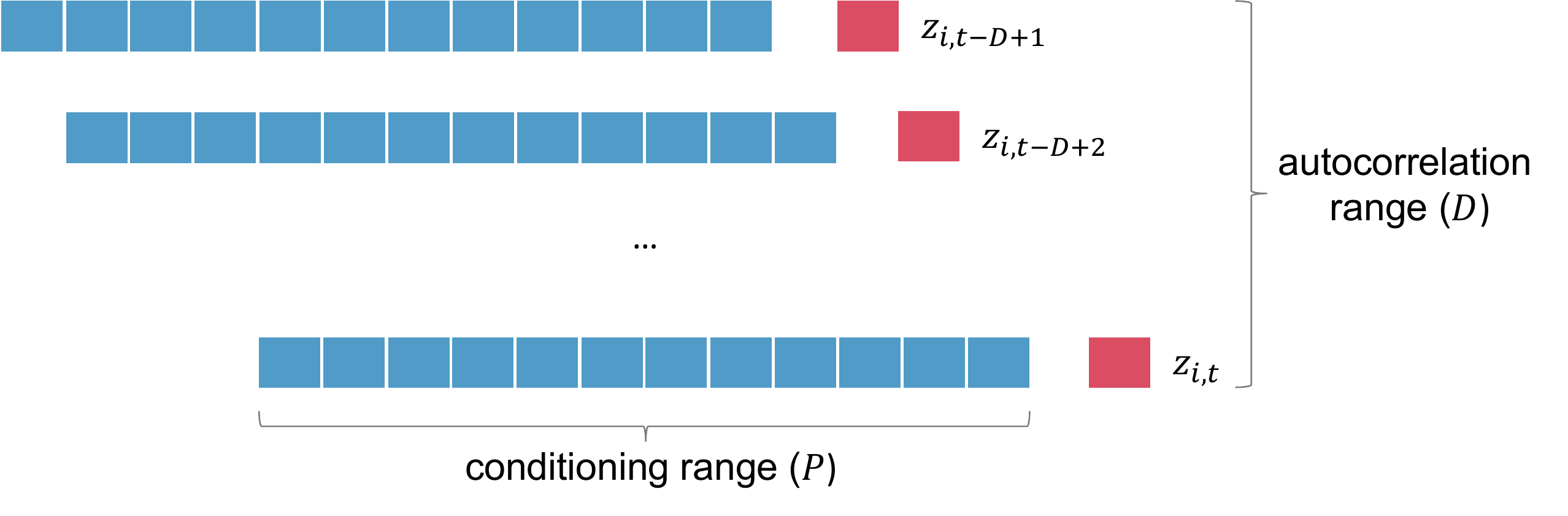}
  \caption{Example of a Mini-batch. Only one-step-ahead prediction is involved during training.}
\label{fig:minibatch}
\end{figure*}

We propose a novel training approach by constructing mini-batches instead of using individual training instances. For most existing deep probabilistic time series models including DeepAR, each training instance consists of a time series segment with a length of $P+Q$, where $P$ represents the conditioning range and $Q$ denotes the prediction range. However, as mentioned, this simple approach cannot characterize the serial correlation of errors among consecutive time steps. To address this issue, we group $D$ consecutive time series segments into a mini-batch, with each segment having a length of $P+1$ (i.e., $Q=1$). In other words, the new training instance (i.e., a mini-batch) becomes a collection of $D$ time series segments with a prediction range $Q=1$. The composition of a mini-batch is illustrated in Fig.~\ref{fig:minibatch}. An example of the collection of target variables in a mini-batch of size $D$ (the time horizon we use for capturing serial correlation) is given by
\begin{equation}
\begin{aligned}
    z_{t-D+1} & = \mu_{t-D+1} + \sigma_{t-D+1}\epsilon_{t-D+1}, \\
    z_{t-D+2} & = \mu_{t-D+2}+ \sigma_{t-D+2} \epsilon_{t-D+2},\\\
    & \ldots\\
    z_{t}& = \mu_{t}+ \sigma_{t}\epsilon_{t}, \\ 
\end{aligned}
\end{equation}
where for time point $t'$, $\mu_{t'}$ and $\sigma_{t'}$ are the output of the model for each time series segment in the mini-batch, and $\epsilon_{t'}$ is the normalized error term. We use boldface symbols to denote the vectors of data and parameters in this mini-batch, e.g.,  $\boldsymbol{z}_t^{\text{bat}}= \left[z_{t-D+1},z_{t-D+2},\ldots,z_{t}\right]^\top$ and the same notation applies to $\boldsymbol{\mu}_t^{\text{bat}}$ and $\boldsymbol{\sigma}_t^{\text{bat}}$. 

Rather than assuming independence among the normalized errors, we consider modeling the joint distribution of the error vector in a mini-batch $\boldsymbol{\epsilon}_t$, denoted as $\boldsymbol{\epsilon}_t^{\text{bat}}= \left[{\epsilon}_{t-D+1},{\epsilon}_{t-D+2},\ldots,{\epsilon}_{t}\right]^\top \sim \mathcal{N} \left( \boldsymbol{0}, \boldsymbol{C}_t \right)$, where $\boldsymbol{C}_t$ is a time-varying correlation matrix. To efficiently characterize the time-varying patterns, we parameterize $\boldsymbol{C}_t$ as a dynamic weighted sum of several base kernel matrices: $\boldsymbol{C}_t = \sum_{m=1}^M w_{m,t}\boldsymbol{K}_m$, where $w_{m,t}\ge 0$ (with $\sum_m w_{m,t}=1$) is the component weight. For each base component, we use a kernel matrix generated from a squared-exponential (SE) kernel function $\boldsymbol{K}^{ij}_m = \exp(-\frac{(i-j)^2}{l_m^2})$ with different lengthscales $l_m$ (e.g., $l_m=1,2,3,\dots$). An identity matrix is included in the additive structure to capture the independent noise process. Taken together, this parameterization ensures that $\boldsymbol{C}_t$ is a positive definite symmetric matrix with unit diagonals, thus being a valid correlation matrix. A small neural network is attached to the original model to project the hidden state to the weights by setting the number of nodes in the final hidden layer to $M$ (i.e., the number of components). A softmax layer is used as the output layer to ensure that these weights sum up to 1. The parameters of the small neural network can be learned jointly with the base model.

The utilization of a time-varying correlation matrix, as opposed to a static correlation matrix, offers the advantage of enabling the model to adapt dynamically to the evolving structure of the error process. For example, the model can assign a higher weight to a kernel matrix generated by a kernel function with a lengthscale of $l=3$ when strong and long-range correlations are present in the current context, whereas it can favor the identity matrix when errors become white noise. This parameterization empowers the model to capture positive autocorrelation that diminishes over time lags. Alternatively, one could opt for a fully learnable positive definite symmetric Toeplitz matrix to parameterize the correlation matrix, which can also accommodate negative and complex correlations. With this formulation, the distribution of $\boldsymbol{\epsilon}_t$ becomes a multivariate Gaussian, which also leads to $\boldsymbol{z}_t^{\text{bat}}\sim \mathcal{N}\left(\boldsymbol{\mu}_t^{\text{bat}}, \boldsymbol{\Sigma}_t^{\text{bat}}\right)$. The $D\times D$ covariance of the associated target variables can be decomposed as $\boldsymbol{\Sigma}_t^{\text{bat}}=\diag(\boldsymbol{\sigma}_t^{\text{bat}})\boldsymbol{C}_t\diag(\boldsymbol{\sigma}_t^{\text{bat}})$. As both $\boldsymbol{\mu}_t^{\text{bat}}$ and $\boldsymbol{\sigma}_t^{\text{bat}}$ are default outputs of the base probabilistic model, the likelihood for a specific time series can be constructed for each mini-batch, and the overall likelihood is given by
\begin{equation}\label{eqn:nll_multi}
\mathcal{L}=\sum_{t=D}^{T} \log p\left(\boldsymbol{z}_{t}^{\text{bat}} \mid \boldsymbol{\mu}_t^{\text{bat}}, \boldsymbol{\Sigma}_t^{\text{bat}}\right).
\end{equation}
By allowing overlap, a total of $T-D+1$ mini-batches can be generated for each time series from the training data. 

\subsection{Multi-step Rolling Prediction}

An autoregressive model performs forecasting in a rolling manner by drawing a sample of the target variable at each time step, feeding it to the next time step as input, and continuing this process until the desired prediction range is reached. Our method can provide extra calibration for this process using the proposed correlation matrix $\boldsymbol{C}_t$. Assume that we have observations till time step $t$ and recall that the collection of normalized errors in a mini-batch jointly follows a multivariate Gaussian distribution. For the next time step $(t+1)$ to be predicted, we have the conditional distribution of $\epsilon_{t+1}$ given the past $(D-1)$ errors using the conditional distribution properties of jointly Gaussian variables:
\begin{multline}\label{eqn:cond_dist}
\epsilon_{t+1} \mid \epsilon_{t}, \epsilon_{t-1},\ldots,\epsilon_{t-D+2} \\ \sim \mathcal{N} \left( \boldsymbol{C}_*\boldsymbol{C}_{\text{obs}}^{-1}\boldsymbol{\epsilon}_{\text{obs}}, 1- \boldsymbol{C}_*\boldsymbol{C}_{\text{obs}}^{-1} \boldsymbol{C}_*^{\top}\right), 
\end{multline}
where $\boldsymbol{\epsilon}_{\text{obs}}=\left[\epsilon_{t-D+2},\ldots,\epsilon_{t-1},\epsilon_t\right]^{\top}\in \mathbb{R}^{D-1}$ represents the set of observed $(D-1)$ residuals at forecasting step $t+1$. Here, $\boldsymbol{C}_{\text{obs}}$ denotes the $(D-1)\times (D-1)$ partition of $\boldsymbol{C}_{t+1}$ that captures the correlations within $\boldsymbol{\epsilon}_{\text{obs}}$, and $\boldsymbol{C}_{*}$ denotes the $1 \times (D-1)$ partition of $\boldsymbol{C}_{t+1}$ that captures the correlations between $\epsilon_{t+1}$ and $\boldsymbol{\epsilon}_{\text{obs}}$, i.e., $\boldsymbol{C}_{t+1}=\begin{bmatrix}
\boldsymbol{C}_{\text{obs}} & \boldsymbol{C}_*^\top \\
\boldsymbol{C}_* & 1 
\end{bmatrix}$, where the weights for generating $\boldsymbol{C}_{t+1}$ are obtained from the hidden state of the base model at time step $t+1$. Note that we remove the time index in $\boldsymbol{C}_{\text{obs}}$, $\boldsymbol{C}_{*}$ and $\boldsymbol{\epsilon}_{\text{obs}}$ for brevity. To obtain a sample of the target variable $\Tilde{z}_{t+1}$, we can first draw a sample of the normalized error $\Tilde{\epsilon}_{t+1}$ using Eq.~\eqref{eqn:cond_dist}. Since both $\mu_{t+1}$ and $\sigma_{t+1}$ are deterministic outcomes from the base model, $\Tilde{z}_{t+1}$ can be derived by
\begin{equation}\label{eqn:sample_tgt}
    \Tilde{z}_{t+1}=\mu_{t+1}+\sigma_{t+1} \Tilde{\epsilon}_{t+1}.
\end{equation}
Based on Eq.~\eqref{eqn:cond_dist} and Eq.~\eqref{eqn:sample_tgt}, it can be seen that the final distribution for ${z}_{t+1}$ becomes
\begin{equation}
\left.z_{t+1}\mid{\bf{h}}_{t+1},\boldsymbol{\epsilon}_{\text{obs}}\right. \sim \mathcal{N}\left(\bar{\mu}_{t+1}, 
\bar{\sigma}_{t+1}^2\right),
\end{equation}
where
\begin{equation}
\begin{aligned}
    & \bar{\mu}_{t+1} = \mu_{t+1} + \sigma_{t+1}\boldsymbol{C}_*\boldsymbol{C}_{\text{obs}}^{-1}\boldsymbol{\epsilon}_{\text{obs}}, \\
    & \bar{\sigma}_{t+1}^2 = \sigma_{t+1}^2\left(1- \boldsymbol{C}_*\boldsymbol{C}_{\text{obs}}^{-1} \boldsymbol{C}_*^{\top}\right).\\
\end{aligned}
\end{equation}

Multi-step rolling prediction can be accomplished by treating the sample $\Tilde{\epsilon}_{t+1}$ as a newly observed residual. Following this process for all subsequent time steps results in a trajectory of $\{\Tilde{\epsilon}_{t+q}\}_{q=1}^Q$.

\section{\MakeUppercase{Experiments}}
\subsection{Datasets and Models}

\begin{table*}[!ht]
\small
\begin{center}
\begin{threeparttable}
\caption{Dataset Summary.}
\label{tab:datasets}
\begin{tabular}{lccccc}
\toprule
Dataset     &   Granularity   &  \# of time series & \# of time steps & $Q$ & Rolling evaluation \\
\midrule
$\mathtt{m4\_hourly}$     &  hourly & 414   & 1,008 & 48 & 1 \\
$\mathtt{exchange\_rate}$     &  workday & 8   & 6,101 & 30 & 5 \\
$\mathtt{m1\_quarterly}$     &  quarterly & 203  & 48 & 8 & 1 \\
$\mathtt{pems03}$     &  5min & 358  & 26,208 & 12 & 24 \\
$\mathtt{pems08}$     &  5min & 170  & 17,856 & 12 & 24 \\
$\mathtt{solar}$     &  hourly & 137  & 7,033 & 24 & 7 \\
$\mathtt{traffic}$       &  hourly & 963   & 4,025 & 24 & 7 \\
$\mathtt{uber\_daily}$      & daily & 262   & 181 & 7 & 1    \\
$\mathtt{uber\_hourly}$           &  hourly  & 262   & 4,344 & 24 & 1   \\
\bottomrule
\end{tabular}
\end{threeparttable}
\end{center}
\end{table*}

We apply the proposed framework to two base prediction models: DeepAR \citep{salinas2020deepar} and an autoregressive decoder-only Transformer (i.e., the GPT model \citep{radford2018improving}). A Gaussian distribution head is employed to generate the distribution parameters for probabilistic forecasting based on the hidden state outputted by the model. It should be noted that our approach can be applied to other autoregressive univariate models without any loss of generality, as long as the final prediction follows a Gaussian distribution. We implemented these models using PyTorch Forecasting \citep{pytorchforecasting}. Both models utilize input data consisting of lagged time series values from the preceding time step, accompanied by supplementary features including time of day, day of the week, and unique time series identifiers. We refer readers to the Supplementary Materials (SM) for comprehensive information on the experiment setup. The code is available at \href{https://github.com/rottenivy/betterbatch}{https://github.com/rottenivy/betterbatch}.

To assess our method's effectiveness, we conducted extensive experiments on diverse real-world time series datasets sourced from GluonTS \citep{alexandrov2020gluonts}. These datasets serve as important benchmarks in evaluating time series forecasting models. The prediction range ($Q$) and the number of rolling evaluations were acquired from each dataset's configuration within GluonTS. Sequential splits into training, validation, and testing sets were performed for each dataset, with the validation set's temporal length matching that of the testing sets. The temporal length of the testing set was computed using the prediction range and the number of rolling evaluations. For example, the testing set for $\mathtt{traffic}$ contains $24+7-1=30$ time steps, indicating 24-step predictions ($Q$) made in a rolling manner across 7 consecutive prediction start timestamps. The details of datasets are summarized in Table~\ref{tab:datasets}. Standardization of the data was carried out using the mean and standard deviation obtained from each time series within the training set. Predictions were subsequently rescaled to their original values for computing evaluation metrics.

\subsection{Evaluation against Baseline}

\begin{table*}[!ht]
\small
\begin{center}
\begin{threeparttable}
\caption{$\operatorname{CRPS}$ Accuracy Comparison.}
\label{tab:crps}
\begin{tabular}{lccrccr}
\toprule
                                 & \multicolumn{3}{c}{DeepAR}              & \multicolumn{3}{c}{Transformer} \\
\cmidrule(lr){2-7}
                                 & w/o                     & w/                         & rel. impr. & w/o                        & w/                         & rel. impr. \\
\midrule
$\mathtt{m4\_hourly}$         & 0.1529$\pm$0.0011          & \textbf{0.1421$\pm$0.0003} & 7.06\%     & 0.1487$\pm$0.0011          & \textbf{0.1431$\pm$0.0009} & 3.77\%     \\
$\mathtt{exchange\_rate}$     & 0.0069$\pm$0.0001          & \textbf{0.0059$\pm$0.0000} & 14.49\%    & 0.0081$\pm$0.0001          & \textbf{0.0074$\pm$0.0001} & 8.64\%     \\
$\mathtt{m1\_quarterly}$      & 0.3767$\pm$0.0006          & \textbf{0.3076$\pm$0.0018} & 18.34\%    & 0.4449$\pm$0.0034          & \textbf{0.3302$\pm$0.0046} & 25.78\%    \\
$\mathtt{pems03}$             & 0.0870$\pm$0.0000          & \textbf{0.0811$\pm$0.0000} & 6.78\%     & 0.0907$\pm$0.0000          & \textbf{0.0845$\pm$0.0001} & 6.84\%     \\
$\mathtt{pems08}$             & 0.0650$\pm$0.0001          & \textbf{0.0588$\pm$0.0000} & 9.54\%     & 0.0622$\pm$0.0000          & \textbf{0.0591$\pm$0.0001} & 4.98\%     \\
$\mathtt{solar}$              & 0.7627$\pm$0.0008          & \textbf{0.7063$\pm$0.0005} & 7.39\%     & 0.6657$\pm$0.0010          & \textbf{0.5379$\pm$0.0010} & 19.20\%    \\
$\mathtt{traffic}$            & 0.2765$\pm$0.0001          & \textbf{0.2387$\pm$0.0001} & 13.67\%    & 0.2422$\pm$0.0002          & \textbf{0.2036$\pm$0.0001} & 15.94\%    \\
$\mathtt{uber\_daily}$        & 0.0897$\pm$0.0003          & \textbf{0.0890$\pm$0.0001} & 0.78\%     & 0.0827$\pm$0.0003          & \textbf{0.0827$\pm$0.0002} & 0.00\%     \\
$\mathtt{uber\_hourly}$       & 0.1549$\pm$0.0007          & \textbf{0.1532$\pm$0.0005} & 1.10\%     & 0.1503$\pm$0.0007          & \textbf{0.1462$\pm$0.0008} & 2.73\%     \\
\cmidrule(lr){1-7}
& & avg. rel. impr. & 8.80\% & & avg. rel. impr. & 9.76\%  \\
\bottomrule
\end{tabular}
\begin{tablenotes}
\item Note: The better results are in boldface (lower is better). All results are based on three runs of each model.
\end{tablenotes}
\end{threeparttable}
\end{center}
\end{table*}

\begin{table*}[!ht]
\small
\begin{center}
\begin{threeparttable}
\caption{Quantile Loss Accuracy Comparison using DeepAR.}
\label{tab:ql}
\begin{tabular}{lccrccr}
\toprule
                              & \multicolumn{3}{c}{$0.5$-risk}   & \multicolumn{3}{c}{$0.9$-risk}   \\
\cmidrule(lr){2-7}
                              & w/o                        & w/                         & rel. impr.   & w/o                        & w/                          & rel. impr.\\
\midrule
$\mathtt{m4\_hourly}$         & 0.1066$\pm$0.0006          & \textbf{0.0999$\pm$0.0006} & 6.29\%       & 0.0579$\pm$0.0006          & \textbf{0.0497$\pm$0.0003}  & 14.16\% \\
$\mathtt{exchange\_rate}$     & 0.0043$\pm$0.0001          & \textbf{0.0042$\pm$0.0000} & 2.33\%       & 0.0028$\pm$0.0000          & \textbf{0.0018$\pm$0.0000}  & 35.71\%\\
$\mathtt{m1\_quarterly}$      & 0.2011$\pm$0.0004          & \textbf{0.1671$\pm$0.0016} & 16.91\%      & 0.3101$\pm$0.0013          & \textbf{0.2489$\pm$0.0030}  & 19.74\% \\
$\mathtt{pems03}$             & 0.0601$\pm$0.0001          & \textbf{0.0563$\pm$0.0000} & 6.32\%       & 0.0337$\pm$0.0000          & \textbf{0.0287$\pm$0.0000}  & 14.84\%\\
$\mathtt{pems08}$             & 0.0456$\pm$0.0001          & \textbf{0.0410$\pm$0.0000} & 10.09\%      & 0.0208$\pm$0.0000          & \textbf{0.0187$\pm$0.0000}  & 10.10\%\\
$\mathtt{solar}$              & 0.5336$\pm$0.0009          & \textbf{0.4896$\pm$0.0005} & 8.25\%       & 0.1827$\pm$0.0001          & \textbf{0.1699$\pm$0.0004}  & 7.01\%\\
$\mathtt{traffic}$            & 0.1608$\pm$0.0000          & \textbf{0.1402$\pm$0.0001} & 12.81\%      & 0.1167$\pm$0.0001          & \textbf{0.1051$\pm$0.0001}  & 9.94\%\\
$\mathtt{uber\_daily}$        & \textbf{0.0598$\pm$0.0006} & 0.0603$\pm$0.0004          & -0.84\%      & 0.0392$\pm$0.0005          & \textbf{0.0369$\pm$0.0002}  & 5.87\% \\
$\mathtt{uber\_hourly}$       & 0.1091$\pm$0.0005          & \textbf{0.1040$\pm$0.0003} & 4.67\%       & \textbf{0.0541$\pm$0.0003} & 0.0574$\pm$0.0003           & -6.10\%\\
\cmidrule(lr){1-7}
& & avg. rel. impr. & 7.42\% & & avg. rel. impr. & 12.36\%  \\
\bottomrule
\end{tabular}
\begin{tablenotes}
\item Note: The better results are in boldface (lower is better). All results are based on three runs of each model.
\end{tablenotes}
\end{threeparttable}
\end{center}
\end{table*}

We evaluate the proposed approach by comparing it with models trained using Gaussian likelihood loss. To simplify the comparison and ensure fairness in terms of the data used during training, we set the autocorrelation range ($D$) to be identical to the prediction range ($Q$). This alignment ensures that each mini-batch in our method covers a time horizon of $P+D$, while in the conventional training method, each training instance spans a time horizon of $P+Q$. By setting $D=Q$, we guarantee that both methods involve the same amount of data per batch, given the same batch sizes. Furthermore, we follow the default configuration in GluonTS by setting the context range equal to the prediction range, i.e., $P=Q$.

In our proposed method, we introduce a small number of additional parameters dedicated to projecting the hidden state into component weights $w_{m,t}$, which play a pivotal role in generating the dynamic correlation matrix $\boldsymbol{C}_t$. In practice, the selection of base kernels should be data-specific, and one should perform residual analysis to determine the most appropriate structure. For simplicity, we use $M=4$ base kernels---three SE kernels with $l=1,2,3$, respectively, and an identity matrix. The different lengthscales capture different decaying rates of autocorrelation. The time-varying component weights can help the model dynamically adjust to different correlation structures observed at different time points. 

We use two different probabilistic scores, namely, the Continuous Ranked Probability Score ($\operatorname{CRPS}$) \citep{gneiting2007strictly} and the quantile loss ($\rho$-risk) \citep{salinas2020deepar}, as the evaluation metrics for uncertainty quantification. $\operatorname{CRPS}$ is defined as
\begin{equation}
    \operatorname{CRPS}(F, y)=\mathbb{E}_F|Y-y|-\frac{1}{2} \mathbb{E}_F\left|Y-Y^{\prime}\right|,
\end{equation}
where $y$ is the observation, $F$ is the cumulative distribution function (CDF) of the target variable $z_t$, $Y$ and $Y^{\prime}$ are independent copies of a set of prediction samples associated with this distribution. We aggregate $\operatorname{CRPS}$ by first summing the $\operatorname{CRPS}$ values across the entire testing horizon for all time series. We then normalize these results by dividing by the sum of the corresponding observations. The quantile loss is defined as
\begin{equation}
    L_\rho\left(Z, \hat{Z}^\rho\right)=2(\hat{Z}^\rho-Z)\left((1-\rho) \mathrm{I}_{\hat{Z}^\rho>Z}-\rho \mathrm{I}_{\hat{Z}^\rho \leq Z}\right),
\end{equation}
where $\mathrm{I}$ is a binary indicator function that equals 1 when the condition is met, $\hat{Z}^\rho$ represents the predicted $\rho$-quantile, and $Z$ represents the ground truth value. The quantile loss serves as a metric to assess the accuracy of a given quantile, denoted by $\rho$, from the predictive distribution. Following \cite{salinas2020deepar}, we summarize the quantile losses for the testing set across all time series segments by evaluating a normalized summation of these losses: $\left(\sum_i L_\rho\left(Z_i, \hat{Z}_i^\rho\right)\right) /\left(\sum_i Z_i\right)$. In this paper, we evaluate the $0.5$-risk and the $0.9$-risk. Comprehensive details regarding the computation of the $\operatorname{CRPS}$ and $\rho$-risk can be found in the SM.

In Table~\ref{tab:crps}, we present a comparative analysis of prediction performance using $\operatorname{CRPS}$. The variants combined with our proposed method (denoted as ``w/'') are compared to their respective original implementations optimized with Gaussian likelihood loss (denoted as ``w/o''). The results demonstrate the effectiveness of our approach in enhancing the performance of both models across a wide range of datasets, yielding notable improvements in predictive accuracy. Our method yields an average improvement of 8.80\% for DeepAR and 9.76\% for Transformer. It should be noted that our method exhibits versatile performance improvements that vary across different datasets. This variability can be attributed to a combination of factors, including the inherent characteristics of the data and the baseline performance of the original model on each specific dataset. Notably, when the original model already achieves exceptional performance on a particular dataset, our method could demonstrate minimal enhancement (e.g., $\mathtt{uber\_daily}$ and $\mathtt{uber\_hourly}$). Moreover, the degree of alignment between our kernel assumption and the true error autocorrelation structure significantly impacts the performance of our method. 

Furthermore, an additional comparison using quantile loss with DeepAR is shown in Table~\ref{tab:ql}, presenting the outcomes for $0.5$-risk and $0.9$-risk. A noteworthy observation is the greater improvement in $0.9$-risk (12.36\%) compared to $0.5$-risk (7.42\%) with our method, implying that the model trained using our approach may provide more reliable prediction intervals for quantifying data uncertainty. Further analyses using other evaluation metrics such as MSE alongside naive baselines are also reported in the SM.

\begin{figure}[!ht]
  \centering
  \includegraphics[width=0.48\textwidth]{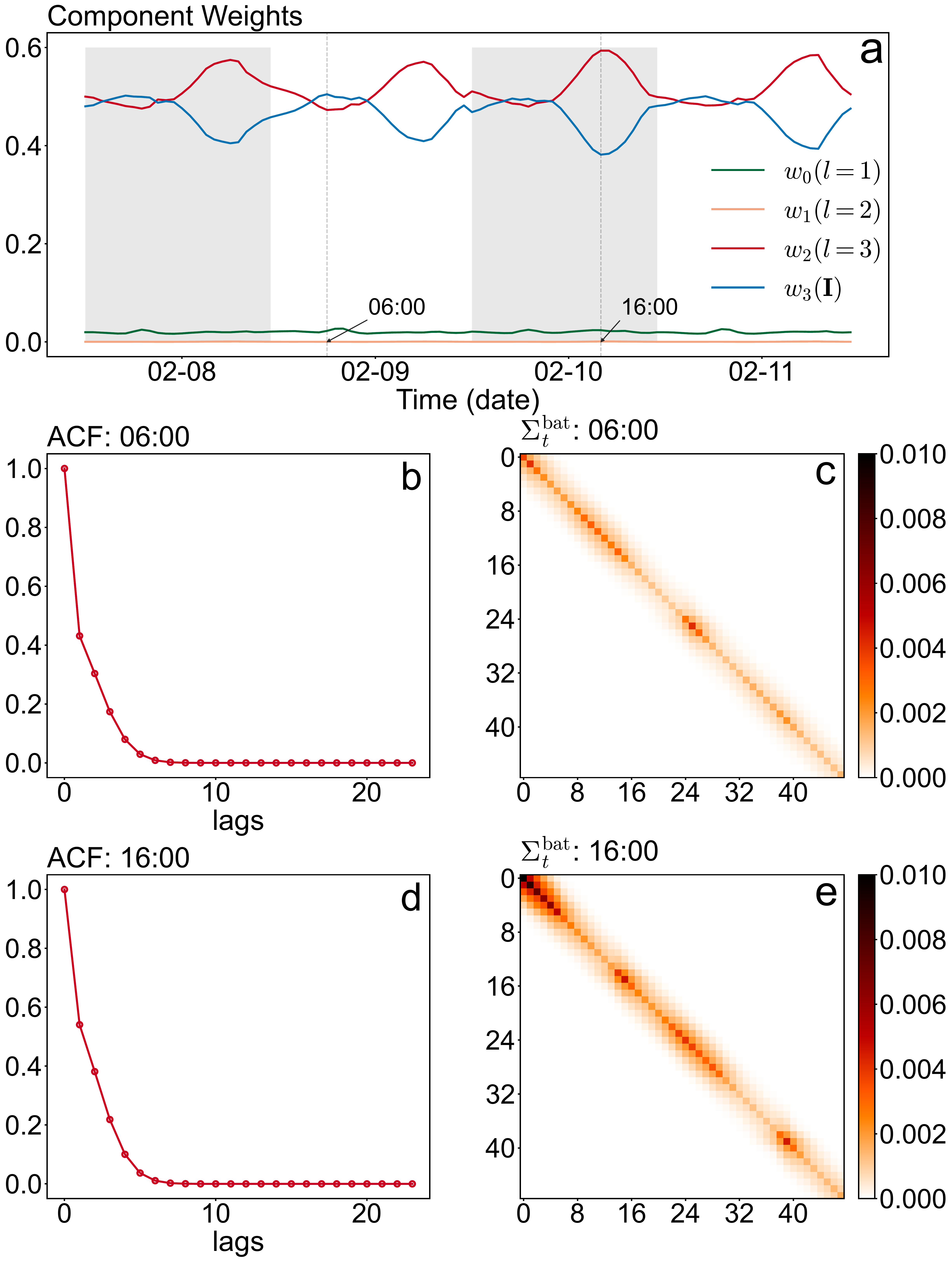}
  \caption{(a) Component weights for generating the correlation matrix of an example time series from the $\mathtt{m4\_hourly}$ dataset. Parameters $w_0, w_1, w_2$ represent the component weights for kernel matrices associated with lengthscales $l=1,2,3$, respectively, and $w_3$ is the component weight for the identity matrix. Shaded areas distinguish different days; (b) and (d): The autocorrelation function (ACF) indicated by the resulting error correlation matrix $\boldsymbol{C}_t$ at 6:00 and 16:00. Given the rapid decay of the ACF, we only plot 24 lags to enhance clarity in our visualization; (c) and (e): The corresponding covariance matrix of the associated target variables $\boldsymbol{\Sigma}_t^{\text{bat}}=\diag(\boldsymbol{\sigma}_t^{\text{bat}})\boldsymbol{C}_t\diag(\boldsymbol{\sigma}_t^{\text{bat}})$ at 6:00 and 16:00, respectively.}
\label{fig:mix_weights}
\end{figure}

\subsection{Interpretation of Correlation}

A key element of our method lies in its ability to capture error autocorrelation through the dynamic construction of a covariance matrix. This is achieved by introducing a dynamic weighted sum of kernel matrices with different lengthscales. The choice of lengthscale significantly influences the structure of autocorrelation---a small lengthscale corresponds to short-range positive autocorrelation, while a large lengthscale can capture positive correlation spanning long lags.

In Fig.~\ref{fig:mix_weights}, we present the generated component weights and the resulting autocorrelation function (i.e., the first row in the learned correlation matrix $\boldsymbol{C}_t$) of an example time series from the $\mathtt{m4\_hourly}$ dataset over a four-day duration. In particular, we observe that the component weights ($w_0, w_1$), corresponding to the kernel matrices with $l=1$ and $l=2$, consistently remain close to zero across the entire observation period. This suggests that the prevailing autocorrelation structure in this dataset is most effectively characterized by the kernel matrix associated with $l=3$. 

Furthermore, we observe the dynamic adjustment of correlation strengths facilitated by the identity matrix over time. Specifically, when $w_3$ (weight assigned to the identity matrix) increases, the error process tends to exhibit greater independence. In contrast, when the weight $w_2$ for the kernel matrix with $l=3$ is larger, the error process becomes more correlated. Figs.~\ref{fig:mix_weights} (b, d) reveal that the autocorrelation at 6:00 in the morning is less pronounced compared to that observed at 16:00. Additionally, Fig.~\ref{fig:mix_weights} (a) demonstrates pronounced daily patterns in autocorrelation, particularly when errors exhibit an increased correlation around 16:00 each day. This underscores the crucial need for our methodology to dynamically adapt the covariance matrix, enabling the effective modeling of these temporal variations. Figs.~\ref{fig:mix_weights} (c, e) depict the covariance matrix of the respective target variables within the autocorrelation horizon. The diagonal elements represent the variance of the target variables generated by the base model, while the off-diagonal elements depict the covariance of the target variables that are facilitated by our approach.

\subsection{Computational Cost}
The proposed method introduces additional computational cost due to the evaluation of the log-likelihood for the multivariate Gaussian distribution in Eq.~\eqref{eqn:nll_multi}. Specifically, this evaluation involves calculating the inverse and determinant of the covariance matrix $\boldsymbol{\Sigma}_t^{\text{bat}}$. However, we can significantly expedite these computations by employing Cholesky decomposition, particularly when the dimensionality $D$ is small. We provide a detailed account of the computational expenses in the SM. Remarkably, our method does not necessarily lead to an increase in training time per epoch or the number of epochs required for convergence. Consequently, the total training time remains comparable to that of conventional training methods in certain scenarios.

\section{\MakeUppercase{Conclusion}}

This paper introduces an innovative training approach to model error autocorrelation in probabilistic time series forecasting. The method involves using mini-batches in training and learning a time-varying covariance matrix that captures the correlation among normalized errors within a mini-batch. Taken together with the standard deviation provided by the base model, we are able to model and predict a time-varying covariance matrix. We implement and evaluate the proposed method using DeepAR and Transformer on various public datasets, and our results confirm the effectiveness of the proposed solution in improving the quality of uncertainty quantification. The broader impact of our method can be observed in two aspects. First, since Gaussian errors are commonly assumed in probabilistic forecasting models, our method can be applied to enhance the training process of such models. Second, the learned autocorrelation can be leveraged to improve multi-step prediction by calibrating the distribution output at each forecasting step. 

There are several directions for future research. First, the kernel-based covariance matrix may be too restrictive for capturing temporal processes. Investigating covariance structures with greater flexibility, such as parameterizing $\boldsymbol{C}_t$ as a fully learnable positive definite symmetric Toeplitz matrix (e.g., AR($p$) process has covariance with a Toeplize structure) and directly factorizing the covariance matrix $\boldsymbol{\Sigma}_t=\boldsymbol{U}_t\boldsymbol{U}_t^{\top}$ (e.g., Wishart process as in \citet{wilson2011generalised}) or the precision matrix $\boldsymbol{\Lambda}_t=\boldsymbol{\Sigma}_t^{-1}=\boldsymbol{V}_t\boldsymbol{V}_t^{\top}$ (e.g., using Cholesky factorization as in \citet{fortuin2020gp}), could be promising avenues. Second, our method can be extended to multivariate models, in which the target output becomes a vector instead of a scalar. A possible solution is to use a matrix Gaussian distribution to replace the multivariate Gaussian distribution used in the current method. This would allow us to learn a full covariance matrix between different target series, thereby capturing any cross-correlations between them.

\subsubsection*{Acknowledgements}
We express our gratitude to all reviewers for their insightful suggestions and comments. We acknowledge the support of the Natural Sciences and Engineering Research Council of Canada (NSERC). Vincent Zhihao Zheng acknowledges the support received from the FRQNT B2X Doctoral Scholarship Program.



\bibliography{ref}




\clearpage
\section*{Checklist}



 \begin{enumerate}

 \item For all models and algorithms presented, check if you include:
 \begin{enumerate}
   \item A clear description of the mathematical setting, assumptions, algorithm, and/or model. [Yes]
   \item An analysis of the properties and complexity (time, space, sample size) of any algorithm. [Yes]
   \item (Optional) Anonymized source code, with specification of all dependencies, including external libraries. [Yes]
 \end{enumerate}

 \item For any theoretical claim, check if you include:
 \begin{enumerate}
   \item Statements of the full set of assumptions of all theoretical results. [Not Applicable]
   \item Complete proofs of all theoretical results. [Not Applicable]
   \item Clear explanations of any assumptions. [Not Applicable]     
 \end{enumerate}

 \item For all figures and tables that present empirical results, check if you include:
 \begin{enumerate}
   \item The code, data, and instructions needed to reproduce the main experimental results (either in the supplemental material or as a URL). [Yes]
   \item All the training details (e.g., data splits, hyperparameters, how they were chosen). [Yes]
         \item A clear definition of the specific measure or statistics and error bars (e.g., with respect to the random seed after running experiments multiple times). [Yes]
         \item A description of the computing infrastructure used. (e.g., type of GPUs, internal cluster, or cloud provider). [Yes]
 \end{enumerate}

 \item If you are using existing assets (e.g., code, data, models) or curating/releasing new assets, check if you include:
 \begin{enumerate}
   \item Citations of the creator If your work uses existing assets. [Yes]
   \item The license information of the assets, if applicable. [Yes]
   \item New assets either in the supplemental material or as a URL, if applicable. [Not Applicable]
   \item Information about consent from data providers/curators. [Not Applicable]
   \item Discussion of sensible content if applicable, e.g., personally identifiable information or offensive content. [Not Applicable]
 \end{enumerate}

 \item If you used crowdsourcing or conducted research with human subjects, check if you include:
 \begin{enumerate}
   \item The full text of instructions given to participants and screenshots. [Not Applicable]
   \item Descriptions of potential participant risks, with links to Institutional Review Board (IRB) approvals if applicable. [Not Applicable]
   \item The estimated hourly wage paid to participants and the total amount spent on participant compensation. [Not Applicable]
 \end{enumerate}

 \end{enumerate}

\end{document}


%
\runningtitle{Better Batch for Deep Probabilistic Time Series Forecasting: Supplementary Materials}

%

\onecolumn
\aistatstitle{Better Batch for Deep Probabilistic Time Series Forecasting: \\
Supplementary Materials}

\section{\MakeUppercase{Dataset Details}}

We conducted experiments on a diverse set of real-world datasets retrieved from GluonTS \citep{alexandrov2020gluonts}, including:

\begin{itemize}
\item $\mathtt{m4\_hourly}$ \citep{makridakis2020m4}: This dataset consists of hourly time series data from various domains, covering microeconomics, macroeconomics, finance, industry, demographics, and other fields. The data originates from the M4-competition.
\item $\mathtt{exchange\_rate}$ \citep{lai2018modeling}: It provides daily exchange rate information for eight different countries spanning the period from 1990 to 2016.
\item $\mathtt{m1\_quarterly}$ \citep{makridakis1982accuracy}: Quarterly time series data from seven different domains.
\item $\mathtt{pems03}$ \citep{chen2001freeway}: Traffic flow records sourced from Caltrans District 3 and acquired through the Caltrans Performance Measurement System (PeMS). The records are aggregated at a 5-minute interval.
\item $\mathtt{pems08}$ \citep{chen2001freeway}: Traffic flow records sourced from Caltrans District 8 and acquired through the Caltrans Performance Measurement System (PeMS). The records are aggregated at a 5-minute interval.
\item $\mathtt{solar}$ \citep{lai2018modeling}: Hourly time series representing solar power production data in the state of Alabama for the year 2006.
\item $\mathtt{traffic}$ \citep{traffic}: Hourly traffic occupancy rate recorded by sensors installed in the San Francisco freeway system between January 2008 and June 2008.
\item $\mathtt{uber\_daily}$ \citep{uber2015tlc}: Daily time series of Uber pickups in New York City from February to July 2015.
\item $\mathtt{uber\_hourly}$ \citep{uber2015tlc}: Hourly time series of Uber pickups in New York City from February to July 2015.
\end{itemize}

These datasets are widely used for benchmarking time series forecasting models. Each dataset follows its default configurations in GluonTS, including granularity and prediction range ($Q$). In our experiments, we set the context range ($P$) to match the prediction range, in line with the default setting in GluonTS. For simplicity, we configure the autocorrelation horizon ($D$) to also equal the prediction range ($Q$). Essentially, in this paper, we have $P=Q=D$.

\section{\MakeUppercase{Continuous Ranked Probability Score}}
Since the target time series variable is assumed to follow a normal distribution, the CRPS can be conveniently calculated as 
\begin{equation}
    \operatorname{CRPS}(F, y)=y(2 F(y)-1)+2 f(y)-\frac{1}{\sqrt{\pi}},\\
\end{equation}
\begin{equation}
    \operatorname{CRPS}\left(F_{\mu, \sigma}, y\right)=\sigma \operatorname{CRPS}\left(F, \frac{y-\mu}{\sigma}\right),
\end{equation}
where $y$ is the observation, for normal distribution, $f(x)=\frac{1}{\sqrt{2\pi}}\exp{(-\frac{x^2}{2})}$, $F(x)=\int_{-\infty}^{x}f(t)dx$, and $F_{\mu, \sigma}(x)=F(\frac{x-\mu}{\sigma})$. $\mu$ and $\sigma$ denote the mean and standard deviation evaluated by 100 samples drawn from the predictive distribution.

\section{\MakeUppercase{Compare with Naive Baselines}}
We compare the proposed methods with two naive baselines: ARIMA \citep{box2015time} and ETS exponential smoothing \citep{hyndman2008forecasting} using a more standard metric, namely mean squared error (MSE):
\begin{equation}
    \mathrm{MSE}=\frac{1}{n} \sum_{i=1}^n\left(Z_i-\hat{Z}_i\right)^2,
\end{equation}
where $\hat{Z}_i$ represents the sample mean for DeepAR and Transformer, computed from 100 samples drawn from the predictive distribution. In contrast, for ARIMA and ETS, the models provide deterministic forecasts directly. We conducted this experiment using the $\mathtt{auto.arima}$ function from the $\mathtt{pmdarima}$ package \citep{pmdarima} and the $\mathtt{ETSModel}$ function from the $\mathtt{statsmodels}$ package \citep{seabold2010statsmodels}. The $\mathtt{auto.arima}$ function can automatically determine the optimal order for an ARIMA model. The training data's temporal length was set to be ten times that of the prediction range. The results are summarized in Table~\ref{tab:arima}. Note that results for datasets with insufficient temporal length for training ARIMA models are not provided.

\begin{table}[!ht]
\small
\begin{center}
\begin{threeparttable}
\caption{MSE Accuracy Comparison.}
\label{tab:arima}
\begin{tabular}{lcccccc}
\toprule
                      & ARIMA     & ETS      & \multicolumn{2}{c}{DeepAR}       &        \multicolumn{2}{c}{Transformer} \\
\cmidrule(lr){2-7}
                       &      &    & w/o              & w/            & w/o              & w/       \\
\midrule
$\mathtt{m4\_hourly}$     & 3416.2127 & 3718.8369 & 1482.3241         & \textbf{1358.1923}   & \textbf{1295.8058}   & 1357.3287           \\
$\mathtt{exchange\_rate}$ & 0.0001  & 0.0001  & 0.0001            & \textbf{0.0001}      & 0.0002               & \textbf{0.0001}     \\
$\mathtt{pems03}$         & 1074.7741 & 1019.2916 & 618.2632          & \textbf{515.2057}    & 647.8133             & \textbf{583.6474}   \\
$\mathtt{pems08}$         & 1093.7592 & 793.8637 & 361.2450          & \textbf{325.1754}    & 356.8268             & \textbf{320.8344}   \\
$\mathtt{solar}$          & 1988.4526 & 1434.8300 & 1637.4979         & \textbf{1323.0383}   & 1339.4904            & \textbf{949.7088}   \\
$\mathtt{traffic}$        & 0.0013   &  0.0021 & 0.0011            & \textbf{0.0008}      & 0.0009               & \textbf{0.0008}     \\
$\mathtt{uber\_daily}$    & 39688.8170 & 19413.8089 & 19366.8854        & \textbf{18370.4863}  & \textbf{15591.5254}  & 15718.5710          \\
$\mathtt{uber\_hourly}$   & 371.6443  & 613.9445 & 64.9663           & \textbf{64.6354}     & 62.1469              & \textbf{60.6224}    \\
\bottomrule
\end{tabular}
\begin{tablenotes}
\item Note: The better results between the ``w/o'' and ``w/'' variants are highlighted in boldface (lower values indicate better performance). All results are based on three runs of each model.
\end{tablenotes}
\end{threeparttable}
\end{center}
\end{table}

\section{\MakeUppercase{Experiment Details}}
\subsection{Model Architecture}
\begin{figure}[!t]
  \centering
  \includegraphics[width=0.99\textwidth, interpolate=false]{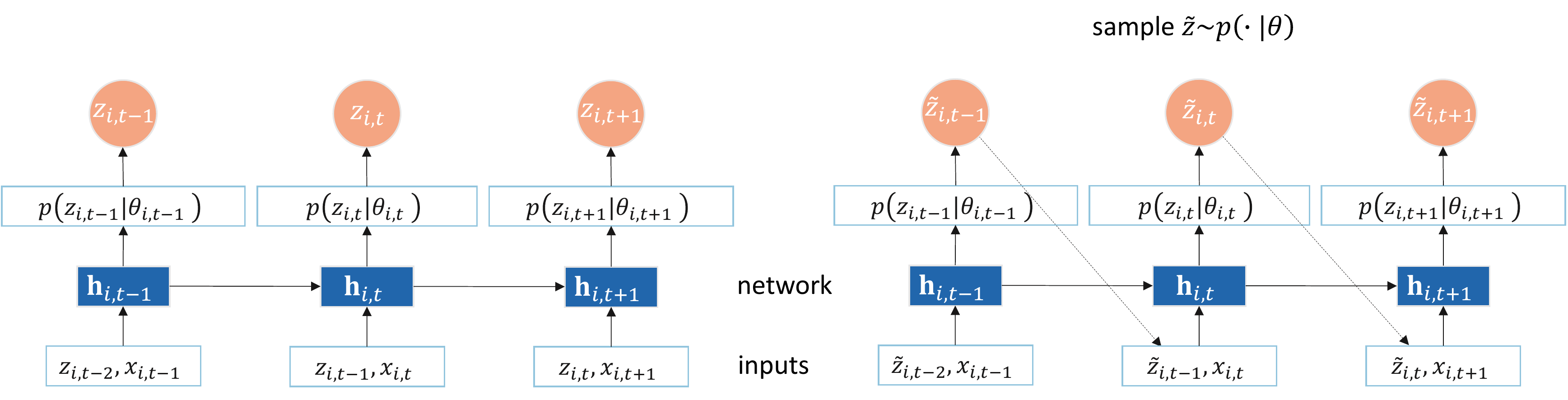}
  \caption{Summary of DeepAR \citep{salinas2020deepar}. Training (left): At each time step \(t\), the inputs to the network are the target value from the previous time step \(z_{i,t-1}\), the covariates \(x_{i,t}\), and the hidden state from previous time step \(\mathbf{h}_{i,t-1}\). The network output \(\mathbf{h}_{i,t}\) is then used to compute the distribution parameters \(\theta_{i,t}=(\mu_{i,t}, \sigma_{i,t})\). For prediction (right), a sample \(\Tilde{z} \sim p(\cdot | \theta_{i,t})\) is drawn and used as input for the next time step in the prediction range, resulting in the generation of a single sample trace. Repeating this prediction process generates multiple traces that collectively represent the joint predictive distribution. Within our framework, the component weights \(\mathbf{w}_{i,t}\) are derived from the same hidden state \(\mathbf{h}_{i,t}\) used for projecting the \(\theta_{i,t}\).}
\label{fig:deepar}
\end{figure}

\begin{figure}[!t]
  \centering
  \includegraphics[width=0.27\textwidth, interpolate=false]{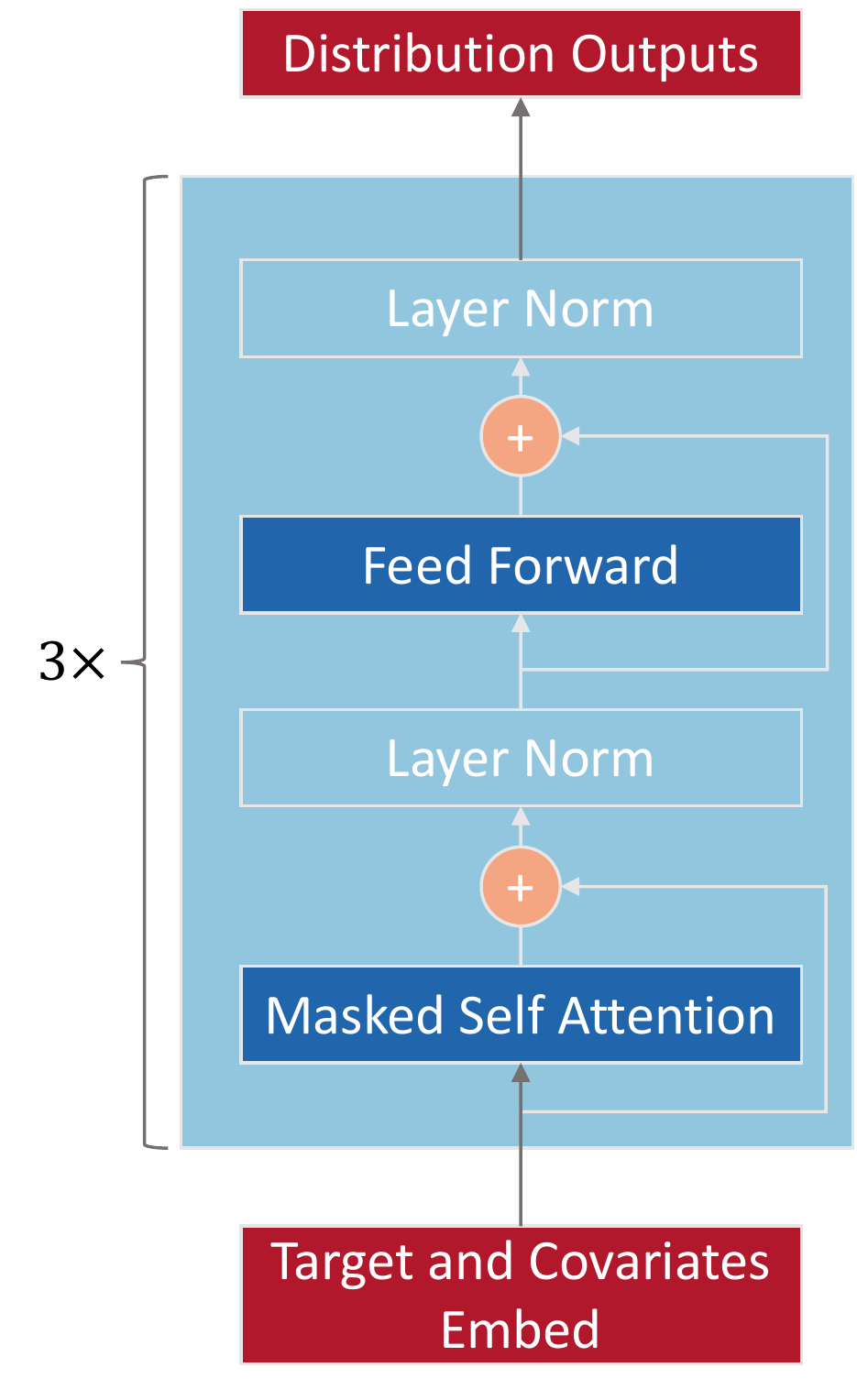}
  \caption{Summary of the Decoder-only Transformer \citep{radford2018improving}. Similarly to DeepAR, Transformer uses ground truth values as inputs during training. However, during the prediction phase, we draw samples and utilize them as inputs for the subsequent time step within the prediction range.}
\label{fig:gpt}
\end{figure}

In this paper, we employ DeepAR \citep{salinas2020deepar} and an autoregressive decoder-only Transformer model, specifically, the GPT model \citep{radford2018improving}, as our base prediction models. The architectures of these two models are visualized in Fig.~\ref{fig:deepar} and Fig.~\ref{fig:gpt}. In the Transformer model, we adapt the output to the parameters of the predictive distribution, namely the mean and standard deviation. To achieve the autoregressive property, we apply a subsequent mask to the input sequence, ensuring that attention scores are computed exclusively based on the inputs preceding the current time step.

\subsection{Features} \label{apx:features}
For datasets featuring hourly granularity, we incorporate two covariates: ``hour of day'' and ``day of week'', as time-varying covariates. In the case of daily datasets, we exclusively employ ``day of week'' as the time-varying covariate. Furthermore, we utilize the identifier of each time series as a static covariate. These covariates are then concatenated with the RNN or Transformer input at each time step post-encoding by the embedding layer. Additionally, to normalize the target values of the time series, we scale them using the mean and standard deviation specific to each time series, obtained from the training dataset.

\subsection{Hyperparameters} \label{hyperparameters}
We employ batch sizes of 64, limiting each epoch to a maximum of 100 batches. The LSTM hyperparameters in DeepAR are adapted from \cite{salinas2020deepar}: comprising three layers of LSTM, each containing 40 nodes, with a dropout rate of 0.1. Additionally, a single linear layer serves as the distribution head to output the distribution parameters. Separate linear layers, along with ELU activation functions, are employed to produce component weights for the base kernel matrices. We configure the Transformer to have a comparable number of parameters with DeepAR. Specifically, we stack three Transformer decoding layers, each with a hidden size of 42, and utilize 2 attention heads ($H=2$). The dropout rate for the Transformer is also set to 0.1.

\subsection{Training Details} \label{apx:training}
Every model underwent training for up to 100 epochs, employing a learning rate of 0.001 with the Adam optimizer. The training process was governed by an early stopping strategy with a patience of 10 epochs. Following the training phase, we restored the best-performing model, determined by validation loss. Details regarding training cost per epoch and the number of epochs until training termination are presented in Table~\ref{tab:training}.

\begin{table}[!ht]
\small
\begin{center}
\begin{threeparttable}
\caption{Training Cost Comparison.}
\label{tab:training}
\begin{tabular}{lcccccccc}
\toprule
                                 & \multicolumn{4}{c}{DeepAR}              & \multicolumn{4}{c}{Transformer} \\
\cmidrule(lr){2-9}
                                 & \multicolumn{2}{c}{w/o}              & \multicolumn{2}{c}{w/}            & \multicolumn{2}{c}{w/o}              & \multicolumn{2}{c}{w/}       \\
                                 & time   & epochs & time & epochs & time        & epochs & time & epochs \\
\midrule
$\mathtt{m4\_hourly}$     & 20     & 99     & 156  & 97     & 20          & 68     & 30   & 95     \\
$\mathtt{exchange\_rate}$ & 17     & 57     & 92   & 33     & 25          & 48     & 28   & 39     \\
$\mathtt{m1\_quarterly}$  & 15     & 10     & 32   & 10     & 28          & 17     & 18   & 18     \\
$\mathtt{pems03}$         & 76     & 65     & 50   & 50     & 84          & 51     & 46   & 45     \\
$\mathtt{pems08}$         & 49     & 25     & 39   & 79     & 68          & 34     & 35   & 99     \\
$\mathtt{solar}$          & 42     & 14     & 53   & 59     & 50          & 51     & 30   & 33     \\
$\mathtt{traffic}$        & 82     & 59     & 55   & 99     & 87          & 59     & 38   & 34     \\
$\mathtt{uber\_daily}$    & 36     & 39     & 14   & 52     & 43          & 48     & 23   & 48     \\
$\mathtt{uber\_hourly}$   & 34     & 46     & 33   & 74     & 38          & 67     & 20   & 98     \\
\bottomrule
\end{tabular}
\begin{tablenotes}
\item Note: ``w/o'' implies the original implementation of models optimized with Gaussian likelihood loss, whereas ``w/'' implies the model combined with our method. ``time'' represents the training time per epoch in seconds.
\end{tablenotes}
\end{threeparttable}
\end{center}
\end{table}

\subsection{Hardware Environment}
Our experiments were conducted under a computer environment with one Intel(R) Xeon(R) CPU E5-2698 v4 @ 2.20GHz and four NVIDIA Tesla V100 GPU. 

\vfill

\bibliography{ref}